\documentclass{article}
\usepackage{spconf,amsmath,graphicx}
\usepackage{comment}

\title{Dialog Context Language Modeling with Recurrent Neural Networks}
\name{{\em Bing Liu$^1$, Ian Lane$^1$$^,$$^2$}}

\address{
  $^1$Electrical and Computer Engineering, Carnegie Mellon University \\
  $^2$Language Technologies Institute, Carnegie Mellon University \\
  {\small \tt liubing@cmu.edu, lane@cmu.edu}
}
\begin{document}
%
\maketitle
\begin{abstract}
In this work, we propose contextual language models that incorporate dialog level discourse information into language modeling. Previous works on contextual language model treat preceding utterances as a sequence of inputs, without considering dialog interactions. We design recurrent neural network (RNN) based contextual language models that specially track the interactions between speakers in a dialog. Experiment results on Switchboard Dialog Act Corpus show that the proposed model outperforms conventional single turn based RNN language model by 3.3\% on perplexity. The proposed models also demonstrate advantageous performance over other competitive contextual language models.

\end{abstract}
\section{Introduction}
\label{sec:intro}
Language model plays an important role in many natural language processing systems, such as in automatic speech recognition \cite{rabiner1993fundamentals,chan2016listen} and machine translation systems \cite{brown1990statistical,cho2014learning}. Recurrent neural network (RNN) based models \cite{mikolov2010recurrent,mikolov2011extensions} have recently shown success in language modeling, outperforming conventional n-gram based models. Long short-term memory \cite{hochreiter1997long,sundermeyer2012lstm} is a widely used RNN variant for language modeling due to its superior performance in capturing longer term dependencies.

Conventional RNN based language model uses a hidden state to represent the summary of the preceding words in a sentence without considering context signals. Mikolov et al. proposed a context dependent RNN language model \cite{mikolov2012context} by connecting a contextual vector to the RNN hidden state. This contextual vector is produced by applying Latent Dirichlet Allocation \cite{blei2003latent} on preceding text. Several other contextual language models were later proposed by using bag-of-word \cite{wang2015larger} and RNN methods \cite{ji2015document} to learn larger context representation that beyond the target sentence. 

The previously proposed contextual language models treat preceding sentences as a sequence of inputs, and they are suitable for document level context modeling. In dialog modeling, however, dialog interactions between speakers play an important role. Modeling utterances in a dialog as a sequence of inputs might not well capture the pauses, turn-taking, and grounding phenomena \cite{clark1991grounding} in a dialog. In this work, we propose contextual RNN language models that specially track the interactions between speakers. We expect such models to generate better representations of the dialog context.

The remainder of the paper is organized as follows. In section 2, we introduce the background on contextual language modeling. In section 3, we describe the proposed dialog context language models. Section 4 discusses the evaluation procedures and results. Section 5 concludes the work.

\section{Background}
\label{sec:background}

\subsection{RNN Language Model}
A language model assigns a probability to a sequence of words $\mathbf{w}=(w_1, w_2, ..., w_{T})$ following probability distribution. 
Using the chain rule, the likelihood of the word sequence $\mathbf{w}$ can be factorized as:
        \begin{equation}
            P(\mathbf{w}) = P(w_1, w_2, ..., w_{T}) = \prod_{t=1}^{T}P(w_{t}|w_{< t}) \\
        \end{equation}
At time step $t$, the system input is the embedding of the word at index $t$, and the system output is the probability distribution of the word at index $t+1$. The RNN hidden state $h_t$ encodes the information of the word sequence up till current step:
        \begin{align}
            &h_t = \operatorname{RNN}(h_{t-1}, w_t) \\
            &P(w_{t+1}|w_{< t+1}) = \operatorname{softmax}(W_{o}h_{t} + b_{o})
        \end{align}
where $W_{o}$ and $b_{o}$ are the output layer weights and biases.

\subsection{Contextual RNN Language Model}
A number of methods have been proposed to introduce contextual information to the language model. Mikolov and Zweig \cite{mikolov2012context} proposed a topic-conditioned RNNLM by introducing a contextual real-valued vector to RNN hidden state. The contextual vector was created by performing LDA \cite{blei2003latent} on preceding text. Wang and Cho \cite{wang2015larger} studied introducing corpus-level discourse information into language modeling. A number of context representation methods were explored, including bag-of-words, sequence of bag-of-words, and sequence of bag-of-words with attention. Lin et al. \cite{lin2015hierarchical} proposed using hierarchical RNN for document modeling. Comparing to using bag-of-words and sequence of bag-of-words for document context representation, using hierarchical RNN can better model the order of words in preceding text, at the cost of the increased computational complexity. These contextual language models focused on contextual information at the document level. Tran et al. \cite{tran2016inter} further proposed a contextual language model that consider information at inter-document level. They claimed that by utilizing the structural information from a tree-structured document set, language modeling performance was largely improved. 

\section{Methods}
\label{sec:methods}

The previously proposed contextual language models focus on applying context by encoding preceding text, without considering interactions in dialogs. These models may not be well suited for dialog language modeling, as they are not designed to capture dialog interactions, such as clarifications and confirmations. By making special design in learning dialog interactions, we expect the models to generate better representations of the dialog context, and thus lower perplexity of the target dialog turn or utterance. 

In this section, we first explain the context dependent RNN language model that operates on utterance or turn level. Following that, we describe the two proposed contextual language models that utilize the dialog level context.

\subsection{Context Dependent RNNLM}
Let $\mathbf{D} = (\mathbf{U}_1, \mathbf{U}_2, ..., \mathbf{U}_K)$ be a dialog that has $K$ turns and involves two speakers. Each turn may have one or more utterances. The $k$th turn $\mathbf{U}_k = (w_1, w_2, ..., w_{T_k})$ is represented as a sequence of $T_k$ words. 
Conditioning on information of the preceding text in the dialog, probability of the target turn $\mathbf{U}_k$ can be calculated as:
        \begin{equation}
            P(\mathbf{U}_k|\mathbf{U}_{<k}) = \prod_{t=1}^{T_k}P(w^{\mathbf{U}_{k}}_{t}|w^{\mathbf{U}_{k}}_{< t}, \mathbf{U}_{<k}) \\
        \end{equation}
where $\mathbf{U}_{<k}$ denotes all previous turns before $\mathbf{U}_k$, and $w^{\mathbf{U}_{k}}_{< t}$ denotes all previous words before the $t$th word in turn $\mathbf{U}_k$.

        \begin{figure}[t]
            \centering
            \includegraphics[width=200pt]{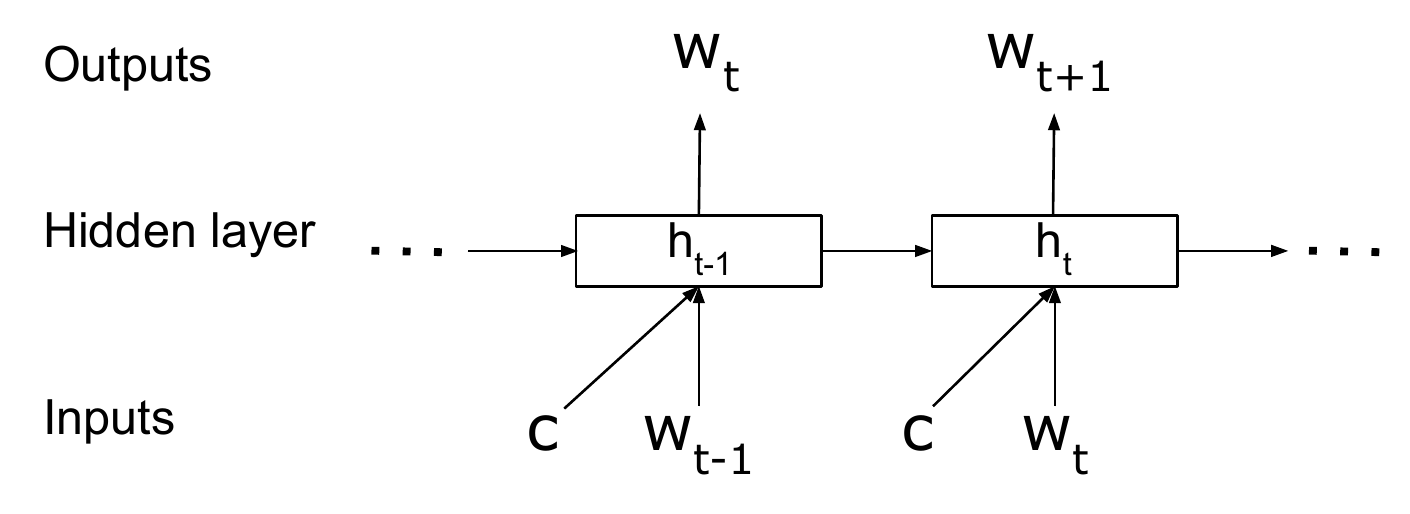}
            \caption{{ Context dependent RNN language model.}}
            \label{fig:Context_RNNLM}
        \end{figure}
In context dependent RNN language model, the context vector $c$ is connected to the RNN hidden state together with the input word embedding at each time step (Figure \ref{fig:Context_RNNLM}). This is similar to the context dependent RNN language model proposed in \cite{mikolov2012context}, other than that the context vector is not connected directly to the RNN output layer. With the additional context vector input $c$, the RNN state $h_t$ is updated as:
        \begin{equation}
            h_t = \operatorname{RNN}(h_{t-1}, [w_t, c])
        \end{equation}

\subsection{Context Representations}
In neural network based language models, the dialog context can be represented as a dense continuous vector. This context vector can be produced in a number of ways. 

One simple approach is to use bag of word embeddings. However, bag of word embedding context representation does not take word order into consideration. An alternative approach is to use an RNN to read the preceding text. The last hidden state of the RNN encoder can be seen as the representation of the text and be used as the context vector for the next turn. To generate document level context representation, one may cascade all sentences in a document by removing the sentence boundaries. The last RNN hidden state of the previous utterance serves as the initial RNN state of the next utterance. As in \cite{ji2015document}, we refer to this model as DRNNLM. Alternatively, in the CCDCLM model proposed in \cite{ji2015document}, the last RNN hidden state of the previous utterance is fed to the RNN hidden state of the target utterance at each time step.

\subsection{Interactive Dialog Context LM}
\begin{figure}[h]
            \centering
            \includegraphics[width=240pt]{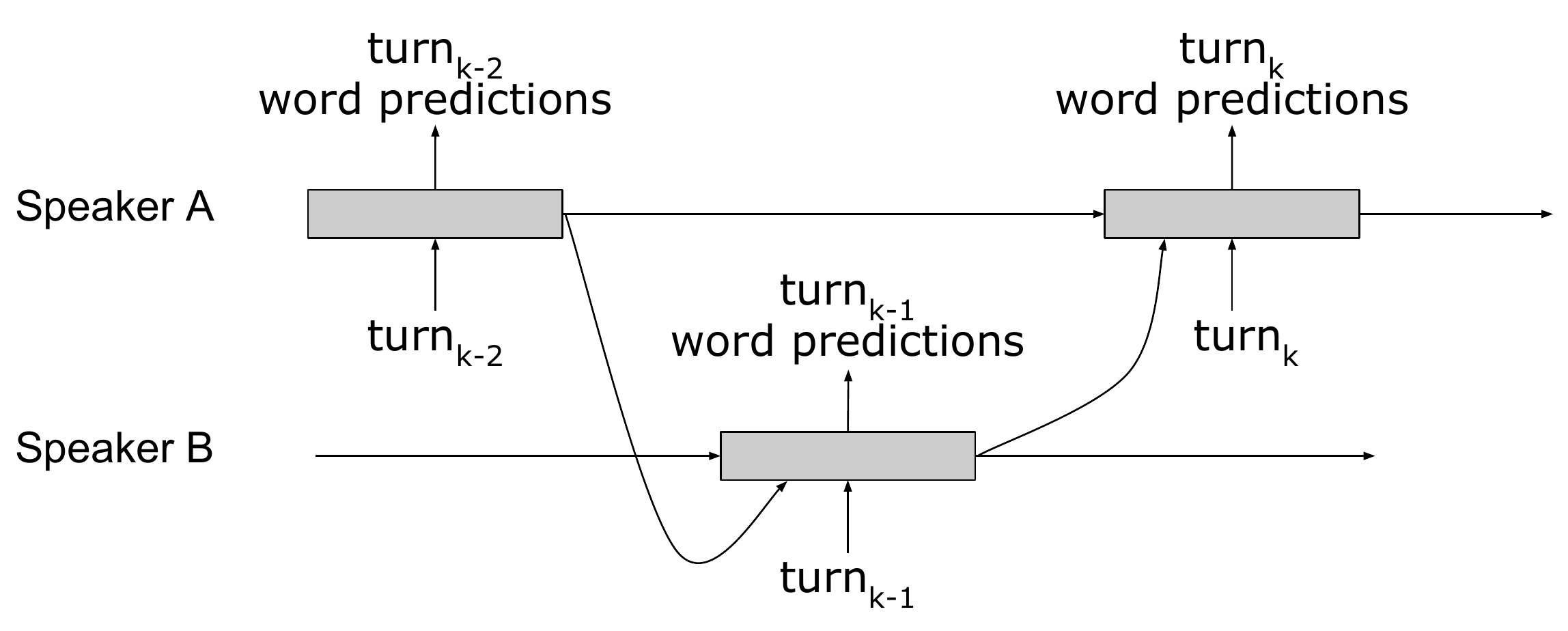}
            \caption{{Interactive Dialog Context Language Model (IDCLM).}}
            \label{fig:IDCLM}
        \end{figure}

The previously proposed contextual language models, such as DRNNLM and CCDCLM, treat dialog history as a sequence of inputs, without modeling dialog interactions. A dialog turn from one speaker may not only be a direct response to the other speaker's query, but also likely to be a continuation of his own previous statement. Thus, when modeling turn $k$ in a dialog, we propose to connect the last RNN state of turn $k-2$ directly to the starting RNN state of turn $k$, instead of letting it to propagate through the RNN for turn $k-1$. The last RNN state of turn $k-1$ serves as the context vector to turn $k$, which is fed to turn $k$'s RNN hidden state at each time step together with the word input. The model architecture is as shown in Figure \ref{fig:IDCLM}. The context vector $c$ and the initial RNN hidden state for the $k$th turn $h^{\mathbf{U}_k}_{0}$ are defined as:
        \begin{equation}
            c = h^{\mathbf{U}_{k-1}}_{T_{k-1}}, \; h^{\mathbf{U}_k}_{0} = h^{\mathbf{U}_{k-2}}_{T_{k-2}}
        \end{equation}
where $h^{\mathbf{U}_{k-1}}_{T_{k-1}}$ represents the last RNN hidden state of turn $k-1$. This model also allows the context signal from previous turns to propagate through the network in fewer steps, which helps to reduce information loss along the propagation. We refer to this model as Interactive Dialog Context Language Model (IDCLM).
\subsection{External State Interactive Dialog Context LM}
\begin{figure}[h]
            \centering
            \includegraphics[width=240pt]{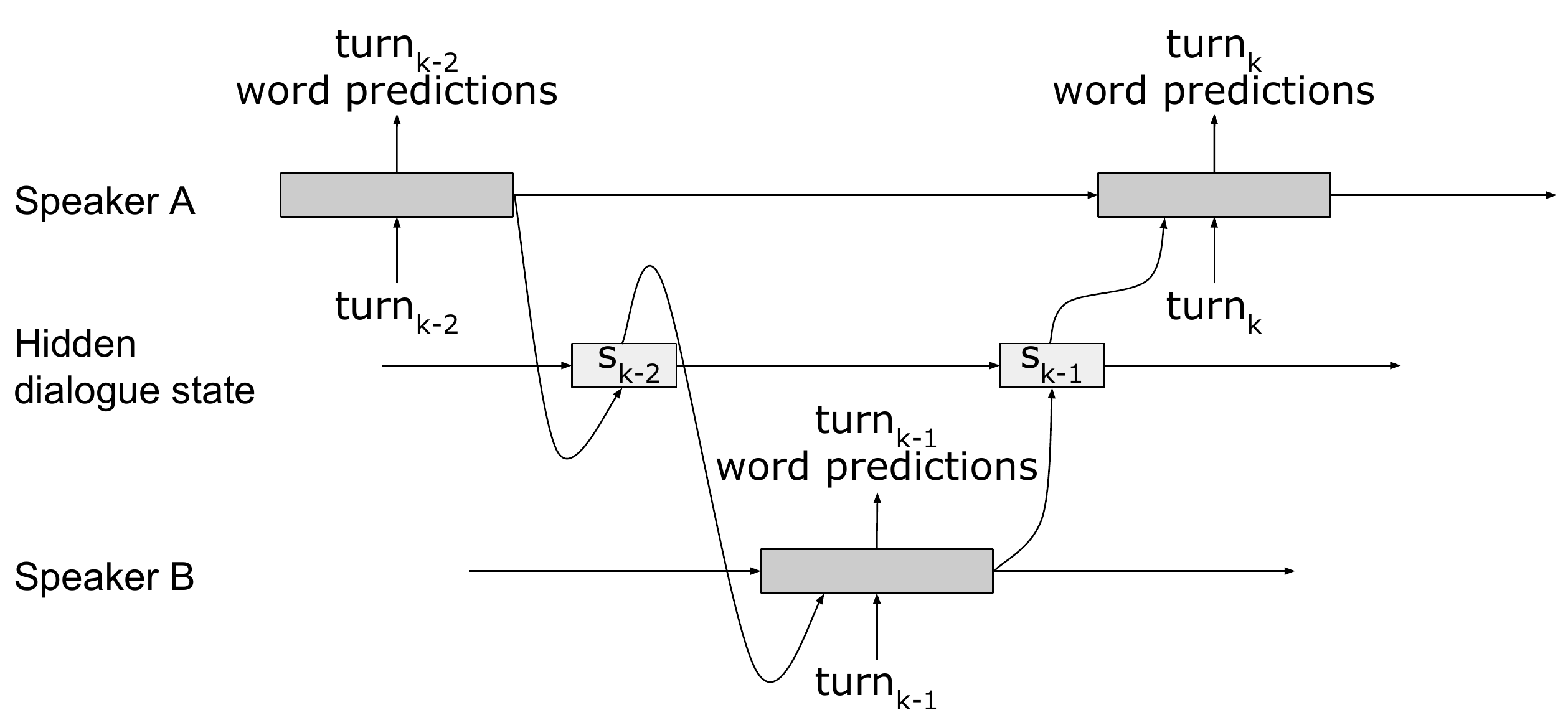}
            \caption{{ External State Interactive Dialog Context Language Model (ESIDCLM). }}
            \label{fig:ESIDCLM}
        \end{figure}

The propagation of dialog context can be seen as a series of updates of a hidden dialog context state along the growing dialog. IDCLM models this hidden dialog context state changes implicitly in the turn level RNN state. Such dialog context state updates can also be modeled in a separated RNN. As shown in the architecture in Figure \ref{fig:ESIDCLM}, we use an external RNN to model the context changes explicitly. Input to the external state RNN is the vector representation of the previous dialog turns. The external state RNN output serves as the dialog context for next turn:
        \begin{equation}
            s_{k-1} = \operatorname{RNN}_{ES}(s_{k-2}, h^{\mathbf{U}_{k-1}}_{T_{k-1}})
        \end{equation}
where $s_{k-1}$ is the output of the external state RNN after the processing of turn $k-1$. The context vector $c$ and the initial RNN hidden state for the $k$th turn $h^{\mathbf{U}_k}_{0}$ are then defined as:
        \begin{equation}
            c = s_{k-1}, \; h^{\mathbf{U}_k}_{0} = h^{\mathbf{U}_{k-2}}_{T_{k-2}}
        \end{equation}
We refer to this model as External State Interactive Dialog Context Language Model (ESIDCLM). 
        
Comparing to IDCLM, ESIDCLM releases the burden of turn level RNN by using an external RNN to model dialog context state changes. One drawback of ESIDCLM is that there are additional RNN model parameters to be learned during model training, which may make the model more prone to overfitting when training data size is limited.

\section{Experiments}
\label{sec:experiments}

\subsection{Data Set}
We use the Switchboard Dialog Act Corpus (SwDA)\footnote{http://compprag.christopherpotts.net/swda.html} in evaluating our contextual langauge models. The SwDA corpus extends the Switchboard-1 Telephone Speech Corpus with turn and utterance-level dialog act tags. The utterances are also tagged with part-of-speech (POS) tags. We split the data in folder sw00 to sw09 as training set, folder sw10 as test set, and folder sw11 to sw13 as validation set. The training, validation, and test sets contain 98.7K turns (190.0K utterances), 5.7K turns (11.3K utterances), and 11.9K turns (22.2K utterances) respectively. Maximum turn length is set to 160. The vocabulary is defined with the top frequent 10K words. 

\subsection{Baselines}
We compare IDCLM and ESIDCLM to several baseline methods, including n-gram based model, single turn RNNLM, and various context dependent RNNLMs. 

\textbf{5-gram KN} \hspace{3mm} A 5-gram language model with modified Kneser-Ney smoothing \cite{chen1996empirical}.

\textbf{Single-Turn-RNNLM} \hspace{3mm} Conventional RNNLM that operates on single turn level with no context information.

\textbf{BoW-Context-RNNLM} \hspace{3mm} Contextual RNNLM with BoW representation of preceding text as context.

\textbf{DRNNLM} \hspace{3mm} Contextual RNNLM with turn level context vector connected to initial RNN state of the target turn.

\textbf{CCDCLM} \hspace{3mm} Contextual RNNLM with turn level context vector connected to RNN hidden state of the target turn at each time step. We implement this model following the design in \cite{ji2015document}.

In order to investigate the potential performance gain that can be achieved by introducing context, we also compare the proposed methods to RNNLMs that use true dialog act tags as context. Although human labeled dialog act might not be the best option for modeling the dialog context state, it provides a reasonable estimation of the best gain that can be achieved by introducing linguistic context. The dialog act sequence is modeled by a separated RNN, similar to the external state RNN used in ESIDCLM. We refer to this model as Dialog Act Context Language Model (DACLM). 

\textbf{DACLM} \hspace{3mm} RNNLM with true dialog act context vector connected to RNN state of the target turn at each time step.

\subsection{Model Configuration and Training}
In this work, we use LSTM cell \cite{hochreiter1997long} as the basic RNN unit for its stronger capability in capturing long-range dependencies in a word sequence comparing to simple RNN. We use pre-trained word vectors~\cite{mikolov2013distributed} that are trained on Google News dataset to initialize the word embeddings. These word embeddings are fine-tuned during model training. We conduct mini-batch training using Adam optimization method following the suggested parameter setup in~\cite{kingma2014adam}. Maximum norm is set to 5 for gradient clipping . For regularization, we apply dropout ($p=0.8$) on the non-recurrent connections~\cite{zaremba2014recurrent} of LSTM. In addition, we apply $L_2$ regularization ($\lambda = 10^{-4}$) on the weights and biases of the RNN output layer.

\subsection{Results and Analysis}
The experiment results on language modeling perplexity for models using different dialog turn size are shown in Table \ref{table:results}. $K$ value indicates the number of turns in the dialog. Perplexity is calculated on the last turn, with preceding turns used as context to the model. 

        \begin{table} [h]
        \caption{\label{table1} {Language modeling perplexities on SwDA corpus with various dialog context turn sizes (K).}}
        \vspace{2mm}
        \centerline{
        \begin{tabular}{l c c c c}
        \hline
        \textbf{Model} & \textbf{K=1} & \textbf{K=2}  & \textbf{K=3}  & \textbf{K=5}\\
        \hline 
        5-gram KN & 65.7 & - & - & -  \\
        Single-Turn-RNNLM & 60.4 & - & - & - \\
        BoW-Context-RNNLM & -    & 59.6 & 59.2 & 58.9   \\
        DRNNLM            & -    & 60.1 & 58.6 & 59.1 \\ 
        CCDCLM            & -    & 63.9 & 61.4 & 62.2 \\ 
        \hline
        IDCLM             & -    & - & 58.8 & 58.6  \\ 
        ESIDCLM           & -    & - & \textbf{58.4} & \textbf{58.5}  \\ 
        \hline
        DACLM  & -    & 58.2 & 57.9 & 58.0  \\ 
        \hline
        \end{tabular}
        \label{table:results}
        }
        \end{table}

As can be seen from the results, all RNN based models outperform the n-gram model by large margin. The BoW-Context-RNNLM and DRNNLM beat the Single-Turn-RNNLM consistently. Our implementation of the context dependent CCDCLM performs worse than Single-Turn-RNNLM. This might due to fact that the target turn word prediction depends too much on the previous turn context vector, which connects directly to the hidden state of current turn RNN at each time step. The model performance on training set might not generalize well during inference given the limited size of the training set. 

The proposed IDCLM and ESIDCLM beat the single turn RNNLM consistently under different context turn sizes. ESIDCLM shows the best language modeling performance under dialog turn size of 3 and 5, outperforming IDCLM by a small margin. IDCLM beats all baseline models when using dialog turn size of 5, and produces slightly worse perplexity than DRNNLM when using dialog turn size of 3.

To analyze the best potential gain that may be achieved by introducing linguistic context, we compare the proposed contextual models to DACLM, the model that uses true dialog act history for dialog context modeling. As shown in Table \ref{table:results}, the gap between our proposed models and DACLM is not wide. This gives a positive hint that the proposed contextual models may implicitly capture the dialog context state changes. 

For fine-grained analyses of the model performance, we further compute the test set perplexity per POS tag and per dialog act tag. We selected the most frequent POS tags and dialog act tags in SwDA corpus, and report the tag based perplexity relative changes ($\%$) of the proposed models comparing to Single-Turn-RNNLM. A negative number indicates performance gain.

        \begin{table} [h]
        \caption{ {Perplexity relative change ($\%$) per POS tag }}
        \vspace{2mm}
        \centerline{
        \begin{tabular}{l c c c}
        \hline
        \textbf{POS Tag} & \textbf{IDCLM} & \textbf{ESIDCLM} & \textbf{DACLM}\\
        \hline 
        PRP & -16.8 & -5.8 & -10.1 \\ 
        IN & -2.0 & -5.5 & -1.8 \\ 
        RB & -4.1 & -8.9 & -4.3 \\ 
        NN & 13.4 & 8.1 & 2.3 \\ 
        UH & -0.4 & 7.7 & -9.7 \\ 
        \hline
        \end{tabular}
        \label{table:pos_tag_results}
        }
        \end{table}

Table \ref{table:pos_tag_results} shows the model perplexity per POS tag. All the three context dependent models produce consistent performance gain over the Single-Turn-RNNLM for pronouns, prepositions, and adverbs, with pronouns having the largest perplexity improvement. However, the proposed contextual models are less effective in capturing nouns. This suggests that the proposed contextual RNN language models exploit the context to achieve superior prediction on certain but not all POS types. Further exploration on the model design is required if we want to better capture words of a specific type.

        \begin{table} [h]
        \caption{ {Perplexity relative change ($\%$) per dialog act tag. }}
        \vspace{2mm}
        \centerline{
        \begin{tabular}{l c c c}
        \hline
        \textbf{DA Tag} & \textbf{IDCLM} & \textbf{ESIDCLM} & \textbf{DACLM}\\
        \hline 
        Statement-non-opinion & -1.8 & -0.5 & -1.6 \\ 
        Acknowledge & -2.6 & 11.4 & -16.3 \\ 
        Statement-opinion & 4.9 & -0.9 & -1.0 \\ 
        Agree/Accept & 14.7 & 2.7 & -15.1 \\ 
        Appreciation & 0.7 & -3.8 & -6.5 \\
        \hline
        \end{tabular}
        \label{table:da_tag_results}
        }
        \end{table}
        
For the dialog act tag based results in Table \ref{table:da_tag_results}, the three contextual models show consistent performance gain on Statement-non-opinion type utterances. The perplexity changes for other dialog act tags vary for different models.

\section{Conclusions}
\label{sec:conclusions}
In this work, we propose two dialog context language models that with special design to model dialog interactions. Our evaluation results on Switchboard Dialog Act Corpus show that the proposed model outperform conventional RNN language model by 3.3\%. The proposed models also illustrate advantageous performance over several competitive contextual language models. Perplexity of the proposed dialog context language models is higher than that of the model using true dialog act tags as context by a small margin. This indicates that the proposed model may implicitly capture the dialog context state for language modeling. 

\bibliographystyle{IEEEbib}
\bibliography{strings,refs}

\end{document}